\documentclass[11pt,a4paper]{article}
\usepackage[hyperref]{emnlp2020}
\usepackage{times}
\usepackage{latexsym}

\usepackage{booktabs}
\usepackage[export]{adjustbox}
\usepackage{pbox}
\usepackage{subcaption}
\usepackage{amsmath}
\usepackage{amssymb}

\usepackage{microtype}

\aclfinalcopy 
\def\aclpaperid{899} 

\title{Supervised Multimodal Bitransformers for Classifying Images and Text}

\author{Douwe Kiela$^\dagger$, Suvrat Bhooshan$^\dagger$, Hamed Firooz$^\dagger$, Ethan Perez$^\ddagger$, Davide Testuggine$^\dagger$\\
$^\dagger$Facebook AI; $^\ddagger$NYU\\
\{dkiela,sbh,mhfirooz,davidet\}@fb.com,perez@nyu.edu
}

\date{}

\begin{document}
\maketitle
\begin{abstract}
Self-supervised bidirectional transformer models such as BERT have led to dramatic improvements in a wide variety of textual classification tasks. The modern digital world is increasingly multimodal, however, and textual information is often accompanied by other modalities such as images. We introduce a simple yet effective baseline for multimodal BERT-like architectures, a \emph{supervised multimodal bitransformer} that jointly finetunes unimodally pretrained text and image encoders by projecting image embeddings to text token space. We approach or match state-of-the-art accuracy on several text-heavy multimodal classification tasks, outperforming strong baselines, including on hard test sets specifically designed to measure multimodal performance. Surprisingly, our method is competitive with ViLBERT, a self-supervised multimodal ``BERT for vision-and-language'' approach, while being much simpler and more easily extendible.
\end{abstract}

\section{Introduction}

Many of the classification problems that we face in the modern digital world are multimodal in nature: textual information on the web rarely occurs alone, and is often accompanied by images, sounds, videos, or other modalities. Recent advances in representation learning for natural language processing, such as BERT \cite{Devlin:2019naacl}, have led to dramatic improvements in text-only classification problems. Following BERT's success, various multimodal architectures have been proposed as well---including ViLBERT~\cite{Lu:2019vilbert}, VisualBERT~\cite{Li:2019visualbert}, LXMERT~\cite{Tan:2019lxmert}, VL-BERT~\cite{Su:2019vlbert} and several others---which advocate pretraining on intermediary or proxy multimodal tasks before finetuning on the multimodal task at hand.

In this work, we describe a simple yet highly effective baseline architecture for BERT-like multimodal architectures. We demonstrate that supervised bidirectional transformers with unimodally pretrained components are excellent at performing multimodal fusion, outperforming a variety of alternative fusion techniques. Moreoever, we find that their performance is competitive with, and can be extended to outperform, multimodally pretrained ViLBERT models on various multimodal classification tasks.

Our proposed approach offers several advantages. Unimodally pretrained models are simpler and easier to adapt to unimodal advances, i.e., it is straightforward to replace the text or image encoders with better alternatives and directly finetune, without requiring multimodal retraining. Furthermore, our method does not rely on a particular feature extraction pipeline since it does not require e.g. region or bounding box proposals, and is modality-agnostic: it works for any sequence of dense vectors. Hence, it can be used to compute raw image features, rather than pre-extracting them, and backpropagate through the entire encoder.



\begin{figure*}[t]
    \centering
    \hspace*{-1.9cm}
    \includegraphics{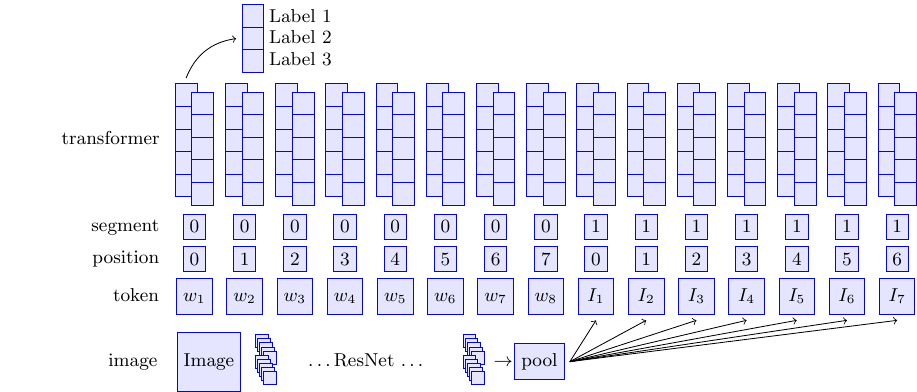}
    \caption{Illustration of the multimodal bitransformer architecture.}
    \label{fig:main}
\end{figure*}

Concretely, our model is BERT-first, learning to map dense multimodal features to BERT's token embedding space. We show that this approach works well on three text-heavy multimodal classification tasks: MM-IMDB~\cite{Arevalo:2017mmimdb}, Food101~\cite{Wang:2015food101} and V-SNLI~\cite{Vu:2018vsnli}. Evaluating on these tasks offers several benefits. Many real-world multimodal tasks on internet data have similar characteristics, in that text is often clearly the dominant modality and the goal is to predict a single classification label rather than answer a question. Importantly, contrary to e.g. VQA \cite{Antol2015vqa}, these tasks have not yet been studied extensively in the multimodal transformer literature. Our work thus allows us to check whether multimodal advances on VQA extend to tasks such as these. Finally, a desired characteristic of multimodal models is improved performance on cases where high-quality multimodal information is available---i.e., the whole should strictly outperform the sum of its parts. We use these tasks to construct novel hard test sets specifically designed to measure the multimodal performance of a system, consisting of examples that unimodal methods fail to classify correctly.




Our findings indicate that the proposed supervised multimodal bitransformer model outperforms various other competitive fusion techniques, 
even if we give those strictly more parameters. We argue that this is due to the multimodal bitransformer's ability to employ self-attention over both modalities simultaneously, providing earlier and more fine-grained multimodal fusion. We find that our straightforward method approaches or matches multimodally pretrained ViLBERT models on our tasks. Put another way, we can match the performance of multimodally pretrained models, without any multimodal pretraining. These results show that the proposed method constitutes a powerful baseline for future work in multimodal classification, as it is straightforward to implement, easy to extend (to different modalities, or different encoders) and performs competitively with more sophisticated methods.

\section{Multimodal Bitransformers}

There is a long history, both in natural language processing and computer vision, of transfer learning from pre-trained representations. Self-supervised word and sentence embeddings \cite{Collobert:2008icml,Mikolov:2013nips,Kiros:2015nips} have become ubiquitous in natural language processing. In computer vision, transferring from supervised ImageNet features is the de facto standard in computer vision \cite{Oquab:2014cvpr,Razavian:2014cvpr}.

While supervised data in NLP has also proven useful for universal sentence representations \cite{Conneau:2017emnlp}, the field was revolutionized by the idea of fine-tuning self-supervised language modeling systems \cite{Dai:2015nips}. Language modeling enables systems to learn embeddings in a contextualized fashion, leading to improved performance on a variety of tasks \cite{Peters:2018naacl,Howard:2018acl}. Training transformers \cite{Vaswani:2017nips} on large quantities of data yielded even better results \cite{Radford:2018tr}. BERT \cite{Devlin:2019naacl} improved on this further by training transformers bidirectionally (which we refer to as bitransformers) and changing the objective to masking, leading to state-of-the-art performance on many tasks.

\begin{table*}[t]
    \centering\small
    \begin{tabular}{llllllll}
        \toprule
        \textbf{Dataset} & \textbf{Source} & \textbf{Type} & \textbf{Train} & \textbf{Dev} & \textbf{Test} & \textbf{\# Inputs} & \textbf{\# Classes}\\
        \midrule
        MM-IMDB & \cite{Arevalo:2017mmimdb} & Multilabel & 15552 & 2608 & 7799 & 2 & 23\\
        FOOD101 & \cite{Wang:2015food101} & Multiclass & 60101 & 5000 & 21695 & 2 & 101\\
        V-SNLI & \cite{Vu:2018vsnli} & Multiclass & 545620 & 9842 & 9842 & 3 & 3\\
        \bottomrule
    \end{tabular}
    \caption{Evaluation tasks used for evaluating performance.}
    \label{tab:dataset_stats}
\end{table*}

We introduce a straightforward yet highly effective multimodal bitransformer model that combines the text-only self-supervised representations from natural language processing with the power of state-of-the-art convolutional neural network architectures from computer vision. See Figure \ref{fig:main} for an illustration of the architecture. In what follows, we describe the different components in more detail.

\subsection{Image Encoder} In computer vision it is common to transfer the final fully connected layer of a pre-trained convolutional neural network \cite{Razavian:2014cvpr}, where the output is often the result of a pooling operation over feature maps. For multimodal bitransformers, however, this pooling is not necessary, since they can handle arbitrary numbers of dense inputs. Thus, we experiment with having the pooling yield not one single output vector, but $N$ separate image embeddings, unlike in a regular convolutional neural network. In this case we use a ResNet-152 \cite{He:2016cvpr} with average pooling over $K\times M$ grids in the image, yielding $N=KM$~output vectors of $2048$ dimensions each, for every image. Images are resized, center-cropped and normalized.

\subsection{Multimodal Transformer Input Layer} We use a bidirectional transformer model initialized with pre-trained BERT weights. The architecture takes contextual embeddings as input, where each contextual embedding is computed as the sum of separate $D$-dimensional segment, position and token embeddings. We learn weights $W_n \in \mathbb{R}^{P \times D}$ to project each of the $N$ image embeddings to $D$-dimensional token input embedding space:

\begin{equation}
    I_n = W_n f(img, n),
\end{equation}

where $f(\cdot, n)$ is the $n$-th output of the image encoder's final pooling operation.

For tasks that consist of a single text and single image input, we assign text one segment ID and image embeddings the other. We use $0$-indexed positional coding, i.e., we start counting from $0$, for each segment. The architecture can be straightforwardly generalized to an arbitrary number of modalities, as we show for the V-SNLI task, which consists of three inputs. Since pre-trained BERT itself has only two segment embeddings, in those cases we initialize additional segment embeddings as $s_i=\frac{1}{2}(s_0+s_1)+\epsilon$ where~$s_i$~is a segment embedding for~$i\geq 2$~and~$\epsilon\sim\mathcal{N}(0,1e^{-2})$. Note that our method is compatible with scenarios where not every modality is present in each example (i.e., if we only have text, or only an image).

\subsection{Classification} We use the first output of the final layer as the input to a classification layer $\mathtt{clf}(x) = W x + b$ where $W\in\mathbb{R}^{D\times C}$, with $D$ as the transformer dimensionality and $C$ as the number of classes. For multilabel tasks, which can have more than one right answer, we apply a sigmoid on the logits and train with a binary cross-entropy loss for each output class (during inference time, we set the threshold at $0.5$); for multiclass tasks we apply a softmax on the logits and train with a regular cross-entropy loss.

\subsection{Pre-training} The image encoder was pre-trained on ImageNet \cite{Deng:2009imagenet}. We use the ResNet-152 \cite{He:2016cvpr} implementation and weights available in PyTorch \cite{Paszke:2017pytorch} through torchvision. We use the pre-trained $12$-layer $768$-dimensional base-uncased model for BERT \cite{Devlin:2019naacl}, trained on English Wikipedia.

\begin{table*}[t]
    \centering
    \small
    \renewcommand\arraystretch{1.1}
    \begin{tabular}{p{2cm}p{1.5cm}p{1.5cm}p{9.5cm}}
        \toprule
        \textbf{Dataset} & \textbf{Label} & \textbf{Image} & \textbf{Text}\\
        \midrule
         MM-IMDB & Comedy & \raisebox{-.9\height}{\includegraphics[width=40pt]{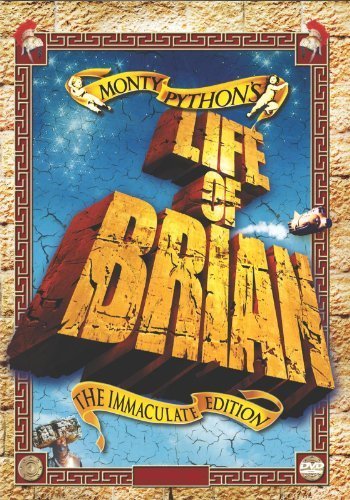}} & Brian is born in a stable on Christmas, right next to You Know Who. The wise men appear and begin to distribute gifts. The star moves further, so they take it all back and move on. This is how Brian's life goes. [...] He joins the Peoples' Front of Judea, one of several dozen separatist groups who actually do nothing, but really hate the Romans. While not about Jesus, it is about those who hadn't time, or interest to listen to his message. Many Political and Social comments.\\\midrule
         FOOD101 & Cup cakes & \raisebox{-.9\height}{\includegraphics[width=40pt]{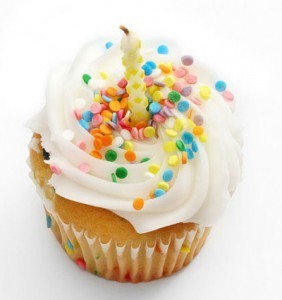}} & [...] simple and oh so delicious these basic cupcakes make a lovely birthday treat makes 24 ingredients 200g unsalted butter softened 1 teaspoon vanilla extract 1 cup caster sugar 3 eggs 2 1 2 cups self raising flour [...] bake for 15 to 17 minutes alternatively for 1 tablespoon capacity mini muffin pans use 1 tablespoon mixture bake for 10 to 12 minutes 4 stand cakes in pans for 2 minutes transfer to a wire rack to cool 5 decorate to suit your party theme [...]\\\midrule
         V-SNLI & Entailment & \raisebox{-.85\height}{\includegraphics[width=40pt]{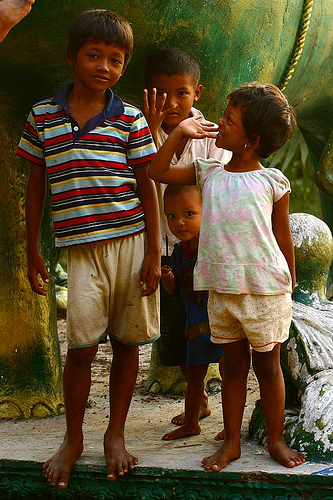}} & \pbox{8cm}{\textbf{Premise}: Children smiling and waving at camera.\\\textbf{Hypothesis}: There are children present.} \\\bottomrule
    \end{tabular}
    \caption{Example data for each of the datasets.}
    \label{tab:examples}
\end{table*}

\subsection{Fine-tuning}

Our architecture consists of a mixture of pre-trained and randomly initialized components. In NLP, BERT is commonly fine-tuned in its entirety, and not transfered as an encoder with fixed parameters, as used to be the case in e.g. SkipThought \cite{Kiros:2015nips} and InferSent \cite{Conneau:2017emnlp}. In computer vision, the convolutional network is often kept fixed \cite{Razavian:2014cvpr}, although it has been found that unfreezing the convolutional network during later stages of training leads to significant improvements, e.g. in image-caption retrieval \cite{Faghri:2017arxiv}.

Multimodal optimization is not trivial \cite{Wang:2019arxiv}. In our model, image embeddings are mapped to BERT's token space using a set of randomly initialized mappings $W_n$. Here, we explore a simple solution for optimization across multiple modalities, namely freezing and unfreezing encoding components at different stages, which we treat as a hyperparameter. If we first learn to map image embeddings to an appropriate subspace of the text encoder's input space, we may expect the network to make more use of visual information than otherwise. Since the text modality is likely to dominate, we want to give the visual modality a chance.

\section{Approach}

In this section, we describe the datasets, the baselines and provide other experimental details.

\subsection{Evaluation}
We evaluate on a diverse set of multimodal classification tasks. We compare against two tasks also used in \cite{Kiela:2018aaai}: MM-IMDB \cite{Arevalo:2017mmimdb} and FOOD101 \cite{Wang:2015food101}. To illustrate that the architecture generalizes beyond two input types, we additionally evaluate on V-SNLI \cite{Vu:2018vsnli}, which consists of (premise, hypothesis, image) triplets. 
See Table \ref{tab:dataset_stats} for dataset statistics and Table \ref{tab:examples} for examples.

\paragraph{MM-IMDB} The MM-IMDB dataset \cite{Arevalo:2017mmimdb} consists of movie plot outlines and movie posters. The objective is to classify each movie by genre. This is a multilabel prediction problem, i.e., one movie can have multiple genres. The dataset was specifically introduced by \cite{Arevalo:2017mmimdb} to address the relative scarcity of high-quality multimodal classification datasets.

\paragraph{FOOD101} The UPMC FOOD101 dataset \cite{Wang:2015food101} contains textual recipe descriptions for 101 food labels. The recipes were scraped from web pages and subsequently cleaned to extract text data. Each page was matched with a single image, where the images were obtained by querying Google Image Search for the given (possibly noisy) category. The objective is to find the corresponding food label for each recipe-image combination.

\paragraph{V-SNLI} The V-SNLI dataset is based on the SNLI dataset \cite{Bowman:2015snli}. The objective is to classify a premise and hypothesis, with associated image, into one of three categories: entailment, neutral or contradition. The SNLI dataset was created by having Turkers provide hypotheses for premises that were derived from captions in the Flickr30k dataset \cite{Young:2014flickr30k}. \cite{Vu:2018vsnli} put the original images and the premise-hypothesis pairs back together in order to create a grounded entailment task, called V-SNLI. V-SNLI also comes with a hard subset of the test set, originally created for SNLI, where a hypothesis-only classifier fails \cite{Gururangan:2018arxiv}.

\subsection{Baselines}

We compare against strong unimodal baselines, as well as the highly competitive, more sophisticated multimodal fusion methods. In all cases we use a single linear classifier, fine-tuning the entire model end-to-end. We describe each of the baselines:

\paragraph{Bag of words (Bow)} We sum 300-dimensional GloVe embeddings \cite{Pennington:2014glove} (Common Crawl) for all words in the text, ignoring the visual features, and feed it to the classifier.
\paragraph{Text-only BERT (Bert)} We take the first output of the final layer of a pre-trained base-uncased BERT model, and feed it to the classifier.
\paragraph{Image-only (Img)} We take a standard pre-trained ResNet-152 with average pooling as output, yielding a $2048$-dimensional vector for each image, and classify it in the same way as the other systems.
\paragraph{Concat Bow + Img (ConcatBow)} We concatenate the outputs of the Bow and the Img baselines. Concatenation is often used as a strong baseline in multimodal methods. In this case, the input to the classifier is $2048$+$300$-dimensions.
\paragraph{Late Fusion} We average the scores of our best Bert and Img classifiers to get the final prediction.
\paragraph{FiLMBert} We combine FiLM \cite{Perez:2018film} with BERT, where the BERT model predicts feature-wise gains and biases for a ConvNet classifier. We use fixed ResNet-152 features as input to the ConvNet, similar to \citet{Perez:2018film}.
\paragraph{Concat BERT + Img (ConcatBert)} We concatenate the outputs of the Bert and the Img baselines. In this case, the input to the classifier is $2048$+$768$-dimensions. This is a competitive baseline, since it combines the best encoder for each modality such that the classifier has direct access to the encoder outputs.

\subsection{Making the Problem Harder}

While we evaluate on a diverse set of multimodal classification tasks, there are actually surprisingly few high-quality tasks of this nature. In many cases, the textual modality is overly dominant \cite[this is even a problem in VQA; see][]{Goyal:2019vqamatter}, making it difficult to tease apart differences between different multimodal methods, or to identify if it is actually worthwhile to incorporate multimodal information in the first place. As we observed earlier, \newcite{Gururangan:2018arxiv} created hard subsets of the SNLI dataset where a hypothesis-only baseline was unable to correctly classify the example, rectifying artifacts in the original SNLI test set. Here, we follow a similar approach, and create hard multimodal test sets for our other two tasks.

\begin{table}[t]
    \centering\small
  \begin{tabular}{lrrrr}
    \toprule
    & MM-IMDB & FOOD-101 & V-SNLI \\\midrule
    GMU & 51.4/63.0 & - & - \\
  	CentralNet & 56.1/63.9 & - & -\\
  	W+V & - & 85.1 & -\\
  	BG & - /62.3 & 90.8 & -\\
  	V-BiMPM & - & - & 86.99\\\midrule
  	Bow & 38.1$\pm$.2/45.6$\pm$.2 & 72.4$\pm$.3 & 48.6$\pm$.3\\
  	Img & 32.5$\pm$.7/44.4$\pm$.3 & 63.2$\pm$.6 & 33.8$\pm$.3\\
  	Bert & 59.9$\pm$.3/65.4$\pm$.1 & 87.2$\pm$.1 & 90.1$\pm$.3\\\midrule
  	Late Fusion & 59.4$\pm$.1/66.2$\pm$.0 & 91.1$\pm$.1 & 90.1$\pm$.0\\
  	ConcatBow & 43.8$\pm$.4/53.6$\pm$.4 & 79.0$\pm$.9 & 49.5$\pm$.1\\
  	FiLMBert & 59.7$\pm$.4/65.1$\pm$.2 & 90.2$\pm$.3 & 90.2$\pm$.2\\
  	ConcatBert & 60.5$\pm$.3/65.9$\pm$.2 & 90.0$\pm$.6 & 90.2$\pm$.4\\
  	MMBT & \textbf{61.6$\pm$.2/66.8$\pm$.1} & \textbf{92.1$\pm$.1} & \textbf{90.4$\pm$.1}\\
  	\bottomrule
  \end{tabular}
    \caption{Main Results. MM-IMDB is Macro F1 / Micro F1; others are Accuracy. Compared against GMU~\cite{Arevalo:2017mmimdb}, CentralNet~\cite{Vielzeuf:2018eccv}, Word2vec+VGGNet (W+V)~\cite{Wang:2015food101}, Bilinear-gated (BG)~\cite{Kiela:2018aaai} and V-BiMPM~\cite{Vu:2018vsnli}.}
    \label{tab:main}
\end{table}

\begin{figure*}[t]
    \centering
\begin{subfigure}{0.4\textwidth}
  \centering
  \includegraphics[width=\linewidth]{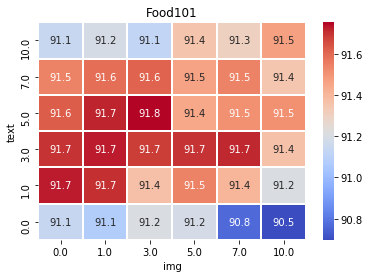}
\end{subfigure}%
\begin{subfigure}{0.4\textwidth}
  \centering
  \includegraphics[width=\linewidth]{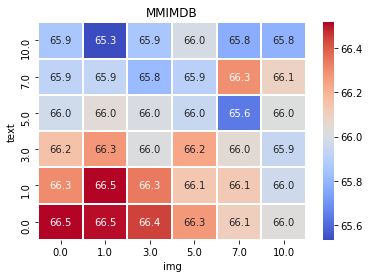}
\end{subfigure}
    \caption{\label{fig:freezing}Analysis of freezing pre-trained text and image components for $N$ epochs of training.}
\end{figure*}

We construct hard test sets by taking the examples where the Bert and Img classifier predictions are most different from the ground truth classes in the test set, i.e. examples that maximize $p(a\neq t|I)p(a\neq t|T)$, where $I$ and $T$ are the image and textual information respectively, $a$ is the predicted answer and $t$ is the correct answer. We take the top 10\% of the most-different examples as the hard cases in the new test sets. The idea is that these are the examples that require more sophisticated multimodal reasoning, allowing us to better examine multimodal-specific performance.


\subsection{Other Implementation Details} For all models, we sweep by over the learning rate (in $\{1e^{-4},5e^{-5}\}$) and early stop on validation accuracy for the multiclass datasets, and Micro-F1 for the multilabel dataset. We additionally sweep over the number of epochs to keep the text and visual encoders fixed, as well as the number of image embeddings to use as input. For the Bert models, we use BertAdam \cite{Devlin:2019naacl} with a warmup rate of $0.1$; for the other models we use regular Adam \cite{Kingma:2014adam}. Since not all datasets are balanced, we weigh the class labels by their inverse frequency. Code and models are available online\footnote{\href{https://github.com/facebookresearch/mmbt}{https://github.com/facebookresearch/mmbt}}.

\section{Results}

The main results can be found in Table \ref{tab:main}. In each case, we show mean performance over $5$ runs with random seeds together with the standard deviation. We compare against the results of \cite{Kiela:2018aaai} on MM-IMDB and FOOD101. They found that a bilinear-gated model worked best, meaning that one of the two input modalities is sigmoided and then gates over the other input bilinearly, i.e. by taking an outer product. Note that in our case, with $2048$-dimensional ResNet outputs and $768$-dimensional Bert outputs, bilinear gated would need a $2048\times768\times101$-dimensional output layer~(approximately 158M parameters just for the classifier on top), which is not practical.

On MM-IMDB, we also compare against Gated Multimodal Units \cite{Arevalo:2017mmimdb}, which are a special recurrent unit specifically designed for multimodal fusion (which similarly has one modality gate over the other). In addition, we compare to CentralNet \cite{Vielzeuf:2018eccv}, a multilayer approach for multimodal fusion that currently holds the state of the art on this dataset. For FOOD101, we include the original results from the paper \cite{Wang:2015food101}, which were obtained by concatenating word2vec and VGGNet features and classifying. For V-SNLI, we compare to the state-of-the-art Visual Bilateral Multi-Perspective Matching (V-BiMPM) model of \cite{Vu:2018vsnli}.

We find that the multimodal bitransformer (MMBT) outperforms the baselines by a significant margin. Late fusion, FiLMBert and ConcatBert perform similarly. We speculate that the cause of MMBT's improvement over ConcatBert is its ability to let information from different modalities interact at different levels, via self-attention, rather than only at the final layer. Part of the improvement comes from Bert's superior performance (which makes sense, given text's dominance), but even then MMBT improves over Bert by e.g.~$\sim$3\%~on MM-IMDB Macro-F1 and an impressive~$\sim$6\%~on Food101 (i.e., an additional ~1300 examples).
In all cases, multimodal models outperform their direct unimodal counterparts.

        

\subsection{Hard Testsets}

Table \ref{tab:hard} reports the results on the hard test sets. Recall that these were created by selecting examples where unimodal (Bert and Img) classifiers differed the most from the ground truth, meaning that these results provide insight into true multimodal performance. We also report results on VSNLI$_{\emph{hard}}$~\cite{Gururangan:2018arxiv}. 

We observe a similar pattern to the main results, with MMBT outperforming the alternatives. Note that on V-SNLI$_{\emph{hard}}$, \newcite{Vu:2018vsnli} report a score of $73.75$ for their best-performing architecture, compared to our $80.4$. It is also interesting to observe that on that hard test set, the image-only classifier already outperforms the text-only one, which is definitely not the case for the normal (non-hard) V-SNLI test set. 


\begin{table}[t]
\centering\small
  \begin{tabular}{lrrrr}
    \toprule
    & MM-IMDB$\dagger$ & FOOD-101$\dagger$ & V-SNLI$\dagger$ \\\midrule
  	Bow & 50.6$\pm$.4 / 54.7$\pm$.4 & 72.7$\pm$.5 & 27.2$\pm$.2\\
  	Img & 39.1$\pm$.9 / 48.2$\pm$.9 & 63.4$\pm$.6 & 32.3$\pm$.3\\
  	Bert & 64.7$\pm$.5 / 67.0$\pm$.3 & 87.3$\pm$.2 & 79.7$\pm$.4\\\midrule
  	Late & 61.7$\pm$.9 / 66.4$\pm$.5 & 91.3$\pm$.5 & 79.6$\pm$.4\\
  	Concat & 64.9$\pm$.4 / 67.2$\pm$.2 & 90.4$\pm$.3 & 79.9$\pm$.9\\
  	MMBT & \textbf{65.3$\pm$.4 / 68.6$\pm$.4} & \textbf{92.4$\pm$.3} & \textbf{80.3$\pm$.1}\\
  	\bottomrule
  \end{tabular}
    \caption{Hard Subsets (marked $\dagger$). Late is Late Fusion. Concat is ConcatBert. MM-IMDB is Macro F1 / Micro F1; others are Accuracy.}
     \label{tab:hard}
\end{table}

\begin{table*}[t]
  \centering\small
  \begin{tabular}{lrrrrr}
    \toprule
    & MM-IMDB & -Hard & FOOD-101 & -Hard\\\midrule
  	MMBT & 61.6$\pm$.2 / 66.8$\pm$.1 & \textbf{65.3$\pm$.4} / \textbf{68.6$\pm$.4} & 92.1$\pm$.1 & \textbf{92.4$\pm$.5}\\ 
	\emph{MMBT-Large} & \emph{63.2$\pm$.2} / \emph{68.0$\pm$.2} & \emph{68.2$\pm$.5} / \emph{70.3$\pm$.4} & \emph{93.2$\pm$.1} & \emph{93.4$\pm$.3}\\\midrule
	ViLBert-VQA & 60.0$\pm$.3 / 66.4$\pm$.2 & 62.7$\pm$.6 / 66.2$\pm$.4 & 92.1$\pm$.1 & 92.4$\pm$.3\\
	ViLBert-VCR & 61.6$\pm$.3 / 67.6$\pm$.2 & 63.4$\pm$.9 / 66.9$\pm$.4 & 92.1$\pm$.1 & 92.1$\pm$.3\\
	ViLBert-Refcoco & 61.4$\pm$.3 / 67.7$\pm$.1 & 63.4$\pm$.5 / 67.1$\pm$.4 & 92.2$\pm$.1 & 92.1$\pm$.3\\
	ViLBert-Flickr30k & 61.4$\pm$.3 / 67.8$\pm$.1 & 63.4$\pm$.9 / 67.0$\pm$.5 & 92.2$\pm$.1 & 92.2$\pm$.3\\
	ViLBert & \textbf{63.0$\pm$.2} / \textbf{68.6$\pm$.1} & \textbf{65.4$\pm$1.} / \textbf{68.6$\pm$.4} & \textbf{92.9$\pm$.1} & \textbf{92.9$\pm$.3}\\
  	\bottomrule
  \end{tabular}
  \caption{Comparison of MMBT to ViLBert on MM-IMDB and FOOD-101.}
  \label{tab:vilbert}
\end{table*}



\subsection{Freezing Strategy}
We conduct an analysis of whether it helps to initially freeze different pre-trained components.
Freezing can help when learning to map from visual space to the expected token input space of the transformer. In other words, the randomly initialized components can be trained first.
We can then unfreeze the image encoder, to make the image information maximally useful, before we unfreeze the bitransformer to tune the entire system on the task. Figure \ref{fig:freezing} shows the results, and indeed corroborates the intuition that it is useful to first learn to put the components together, then unfreeze the image encoder, and only after that unfreeze the pre-trained bitransformer. The optimal number of epochs is task-dependent, while unfreezing the image encoder early works best.

\subsection{Number of Parameters} A possible explanation for the superior performance of the multimodal bitransformer over ConcatBert could be that it has slightly more parameters (i.e., an additional $2048\times D$ versus $2048\times N$, where $D$ is the embedding dimensionality and $N$ is the number of classes), although the difference is small: 168M vs 170M parameters. To investigate this, we also compare against a ConcatBert with a $2$-layer and $3$-layer multi-layer perceptron (MLP) classifier on top, of 174M and 175M parameters respectively, rather than the single-layer logistic regression in MMBT. For MM-IMDB, ConcatBert-2 and ConcatBert-3 get a Macro-F1 of~$60.21\pm.5$~and~$59.71\pm.4$ and a Micro-F1 of ~$65.08\pm.3$~and $64.82\pm.2$~respectively; while for Food101 they get~$91.13\pm.2$~and~$90.27\pm.2$. This clearly demonstrates (cf. Table \ref{tab:main}) that MMBT is superior to ConcatBert, even when we give an already highly competitive baseline even more parameters and a deeper classifier. The results suggest that ConcatBert is more prone to overfitting\footnote{The result was the same with more image embeddings.}.




\begin{figure}[t]
    \centering
    \includegraphics[width=.4\textwidth]{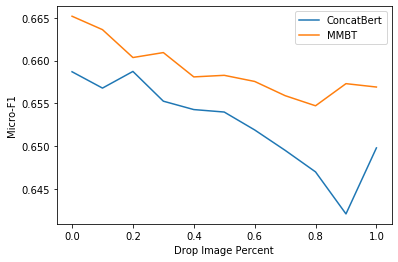}
    \caption{Performance (MicroF1) on MM-IMDB when we drop the image for a percentage of the training set, measuring robustness to missing images.}
    \label{fig:dropimg}
\end{figure}

\subsection{Robustness to Missing Modalities} 

We compare ConcatBert and MMBT in a setting where only a subset of the dataset has images. To our knowledge, this setting has not been explored thoroughly in the literature. It is unclear a priori which of the two models would be more robust to this data regime, and this experiment provides a useful extra dimension for comparing mid-level fusion with the more sophisticated type of fusion provided by MMBT. Figure \ref{fig:dropimg} shows that performance drops with fewer images. 
It is interesting to observe that MMBT is much more robust to missing images than ConcatBert.

\subsection{Comparison to ViLBERT}

We examine the effectiveness of fusing unimodally pretrained components by comparing to self-supervised multimodally pretrained models. We take ViLBERT \cite{Lu:2019vilbert} as the canonical example of that class of models. ViLBERT was trained multimodally on images and captions, and is meant to be the ``BERT of vision and language''. It uses Faster RCNN-extracted bounding boxes
, kept fixed during training. Our focus on these somewhat out-of-the-ordinary tasks now proves fruitful, since it allows us to compare these models on a level playing field.

Table \ref{tab:vilbert} shows the results. We compare against a variety of ViLBert models, both the standard pre-trained version as well as the versions fine-tuned for particular tasks like VQA. The latter approach is not proposed in the original ViLBert paper, but similar ``two-stage pre-training'' approaches have proven effective for fine-tuning BERT on unimodal tasks~\cite{Phang2018sentence}.
We tune using the hyperparameter sets used in that paper: (batch size, learning rate) $\in \{(64, 2e^{-5}), (256, 4e^{-5})\}$.
We observe that our straightforward MMBT model is surprisingly competitive. On MM-IMDB, it matches the task-specific ViLBERT models on Macro-F1. On the Hard subset of that dataset, which more accurately measures multimodal performance, MMBT matches ViLBert's performance. For FOOD-101, we observe a similar story, with performance being remarkably close, occasionally outperforming task-specific models, in particular on the Hard subset. Our results suggest that self-supervised multimodal pre-training has more room for improvement, and that the supervised fusion of unimodally-pretrained components is remarkably competitive.

Our method may be more preferable depending on the constraints: with new models coming out every month, these will be easy to incorporate into the architecture. To illustrate this point~(obviously not a fair comparison), we use a BERT-Large model instead to make MMBT outperform ViLBERT. This is trivial to do in our setting, but for ViLBERT would require retraining from scratch.

\section{Related Work}

Transformers~\cite{Vaswani:2017nips} have been used to encode sequential data for classification with great success when pre-trained for language modeling or language masking and subsequently fine-tuned~\cite{Radford:2018tr,Devlin:2019naacl}.

The question of how to effectively do multimodal fusion has a long history~\cite{Baltrusaitis:2019survey}. While concatenation can be considered the default, other fusion methods have been explored e.g. for lexical representation learning~\cite{Bruni:2014jair,Lazaridou:2015skipgram}. In classification,~\newcite{Kiela:2018aaai} examine various fusion methods for pre-trained fixed representations, and find that a bilinear combination of data with gating worked best. Our supervised multimodal bitransformer has fusion between the modalities via self-attention over many different layers.

Applications in multimodal NLP range from classification to cross-modal retrieval~\cite{Weston:2011wsabie,Frome:2013devise,Socher:2013nips} to image captioning~\cite{Bernardi:2016captioning} to visual question answering~\cite{Antol2015vqa} and multimodal machine translation~\cite{Elliott:2017emnlp}. Multimodal information is also useful in learning human-like meaning representations~\cite{Baroni:16,Kiela:17thesis}. Multimodal bitransformers provide what is effectively a deep fusion method. Related deep fusion methods include multimodal transformers~\cite{Tsai:2019acl}, CentralNet~\cite{Vielzeuf:2018eccv}, MFAS~\cite{Perez-Rua:2019arxiv} and Tensor Fusion Networks~\cite{Zadeh2017tfn}.

There has been a large number of self-supervised multimodal architectures published recently, e.g. ViLBERT~\cite{Lu:2019vilbert}, VisualBERT~\cite{Li:2019visualbert}, LXMERT~\cite{Tan:2019lxmert}, VL-BERT~\cite{Su:2019vlbert}, VideoBERT~\cite{Sun:2019arxiv}, and others. Our model differs from these self-supervised architectures in that the individual components are pretrained only unimodally. This has pros and cons: our method is straightforward and intuitive, easy to implement even for existing self-supervised encoders, and obtains impressive improvements. If a new and better text or vision model comes out, it is trivial to replace components. On the other hand, it is not able to fully leverage multimodal information during self-supervised pre-training. That said, it does potentially have access to orders of magnitude more unimodal data. In other words, if anything, these supervised multimodal bitransformers should provide a strong baseline for gauging if and how much self-supervised multimodal pretraining actually helps.

\section{Conclusion}

In this work, we introduced a supervised multimodal bitransformer model. We compared against several baselines on a variety of tasks, including on hard test sets created specifically for examining multimodal performance (i.e., where unimodal performance fails). We find that the proposed architecture significantly outperforms the existing state of the art, as well as strong baselines. We then conducted an analysis of multimodal optimization, exploring a freezing/unfreezing strategy, and looked at the number of parameters, showing that the strong baseline with more parameters and a deeper classifier was still outperformed.

Our architecture consists of components that were pre-trained individually as unimodal tasks, which already showed great improvements over alternatives. It is as of yet unclear if multimodal self-supervised models are going to be generally useful. We compared to ViLBERT and showed that the proposed model performs competitively, while being much simpler. The methods outlined here should serve as a useful and powerful baseline to gauge the performance of self-supervised multimodal models. Supervised multimodal bitransformers are straightforward and intuitive, and importantly, are easy to implement even for existing self-supervised encoders.

\bibliography{anthology,emnlp2020}

\begin{thebibliography}{49}
\expandafter\ifx\csname natexlab\endcsname\relax\def\natexlab#1{#1}\fi

\bibitem[{Antol et~al.(2015)Antol, Agrawal, Lu, Mitchell, Batra,
  Lawrence~Zitnick, and Parikh}]{Antol2015vqa}
Stanislaw Antol, Aishwarya Agrawal, Jiasen Lu, Margaret Mitchell, Dhruv Batra,
  C~Lawrence~Zitnick, and Devi Parikh. 2015.
\newblock Vqa: Visual question answering.
\newblock In \emph{Proceedings of the IEEE international conference on computer
  vision}, pages 2425--2433.

\bibitem[{Arevalo et~al.(2017)Arevalo, Solorio, Montes-y G{\'o}mez, and
  Gonz{\'a}lez}]{Arevalo:2017mmimdb}
John Arevalo, Thamar Solorio, Manuel Montes-y G{\'o}mez, and Fabio~A
  Gonz{\'a}lez. 2017.
\newblock Gated multimodal units for information fusion.
\newblock \emph{arXiv preprint arXiv:1702.01992}.

\bibitem[{Baltru{\v{s}}aitis et~al.(2019)Baltru{\v{s}}aitis, Ahuja, and
  Morency}]{Baltrusaitis:2019survey}
Tadas Baltru{\v{s}}aitis, Chaitanya Ahuja, and Louis-Philippe Morency. 2019.
\newblock Multimodal machine learning: A survey and taxonomy.
\newblock \emph{IEEE Transactions on Pattern Analysis and Machine
  Intelligence}, 41(2):423--443.

\bibitem[{Baroni(2016)}]{Baroni:16}
Marco Baroni. 2016.
\newblock Grounding distributional semantics in the visual world.
\newblock \emph{Language and Linguistics Compass}, 10(1):3--13.

\bibitem[{Bernardi et~al.(2016)Bernardi, Cakici, Elliott, Erdem, Erdem,
  Ikizler-Cinbis, Keller, Muscat, and Plank}]{Bernardi:2016captioning}
Raffaella Bernardi, Ruket Cakici, Desmond Elliott, Aykut Erdem, Erkut Erdem,
  Nazli Ikizler-Cinbis, Frank Keller, Adrian Muscat, and Barbara Plank. 2016.
\newblock Automatic description generation from images: A survey of models,
  datasets, and evaluation measures.
\newblock \emph{Journal of Artificial Intelligence Research}, 55:409--442.

\bibitem[{Bowman et~al.(2015)Bowman, Angeli, Potts, and
  Manning}]{Bowman:2015snli}
Samuel~R Bowman, Gabor Angeli, Christopher Potts, and Christopher~D Manning.
  2015.
\newblock A large annotated corpus for learning natural language inference.
\newblock \emph{arXiv preprint arXiv:1508.05326}.

\bibitem[{Bruni et~al.(2014)Bruni, Tran, and Baroni}]{Bruni:2014jair}
Elia Bruni, Nam-Khanh Tran, and Marco Baroni. 2014.
\newblock Multimodal distributional semantics.
\newblock \emph{Journal of Artificial Intelligence Research}, 49:1--47.

\bibitem[{Collobert and Weston(2008)}]{Collobert:2008icml}
Ronan Collobert and Jason Weston. 2008.
\newblock A unified architecture for natural language processing: Deep neural
  networks with multitask learning.
\newblock In \emph{Proceedings of the 25th international conference on Machine
  learning}, pages 160--167. ACM.

\bibitem[{Conneau et~al.(2017)Conneau, Kiela, Schwenk, Barrault, and
  Bordes}]{Conneau:2017emnlp}
Alexis Conneau, Douwe Kiela, Holger Schwenk, Loic Barrault, and Antoine Bordes.
  2017.
\newblock Supervised learning of universal sentence representations from
  natural language inference data.
\newblock \emph{arXiv preprint arXiv:1705.02364}.

\bibitem[{Dai and Le(2015)}]{Dai:2015nips}
Andrew~M Dai and Quoc~V Le. 2015.
\newblock Semi-supervised sequence learning.
\newblock In \emph{Advances in neural information processing systems}, pages
  3079--3087.

\bibitem[{Deng et~al.(2009)Deng, Dong, Socher, Li, Li, and
  Fei-Fei}]{Deng:2009imagenet}
Jia Deng, Wei Dong, Richard Socher, Li-Jia Li, Kai Li, and Li~Fei-Fei. 2009.
\newblock Imagenet: A large-scale hierarchical image database.
\newblock In \emph{2009 IEEE conference on computer vision and pattern
  recognition}, pages 248--255. Ieee.

\bibitem[{Devlin et~al.(2019)Devlin, Chang, Lee, and
  Toutanova}]{Devlin:2019naacl}
Jacob Devlin, Ming-Wei Chang, Kenton Lee, and Kristina Toutanova. 2019.
\newblock Bert: Pre-training of deep bidirectional transformers for language
  understanding.
\newblock \emph{Proceedings of NAACL}.

\bibitem[{Elliott et~al.(2017)Elliott, Frank, Barrault, Bougares, and
  Specia}]{Elliott:2017emnlp}
Desmond Elliott, Stella Frank, Lo\"{i}c Barrault, Fethi Bougares, and Lucia
  Specia. 2017.
\newblock Findings of the second shared task on multimodal machine translation
  and multilingual image description.
\newblock In \emph{Proceedings of the Second Conference on Machine Translation,
  Volume 2: Shared Task Papers}, pages 215--233, Copenhagen, Denmark.

\bibitem[{Faghri et~al.(2017)Faghri, Fleet, Kiros, and
  Fidler}]{Faghri:2017arxiv}
Fartash Faghri, David~J Fleet, Jamie~Ryan Kiros, and Sanja Fidler. 2017.
\newblock Vse++: Improving visual-semantic embeddings with hard negatives.
\newblock \emph{arXiv preprint arXiv:1707.05612}.

\bibitem[{Frome et~al.(2013)Frome, Corrado, Shlens, Bengio, Dean, Mikolov
  et~al.}]{Frome:2013devise}
Andrea Frome, Greg~S Corrado, Jon Shlens, Samy Bengio, Jeff Dean, Tomas
  Mikolov, et~al. 2013.
\newblock Devise: A deep visual-semantic embedding model.
\newblock In \emph{Advances in neural information processing systems}, pages
  2121--2129.

\bibitem[{Goyal et~al.(2019)Goyal, Khot, Agrawal, Summers-Stay, Batra, and
  Parikh}]{Goyal:2019vqamatter}
Yash Goyal, Tejas Khot, Aishwarya Agrawal, Douglas Summers-Stay, Dhruv Batra,
  and Devi Parikh. 2019.
\newblock Making the v in vqa matter: Elevating the role of image understanding
  in visual question answering.
\newblock \emph{Int. J. Comput. Vision}, 127(4):398–414.

\bibitem[{Gururangan et~al.(2018)Gururangan, Swayamdipta, Levy, Schwartz,
  Bowman, and Smith}]{Gururangan:2018arxiv}
Suchin Gururangan, Swabha Swayamdipta, Omer Levy, Roy Schwartz, Samuel~R
  Bowman, and Noah~A Smith. 2018.
\newblock Annotation artifacts in natural language inference data.
\newblock \emph{arXiv preprint arXiv:1803.02324}.

\bibitem[{He et~al.(2016)He, Zhang, Ren, and Sun}]{He:2016cvpr}
Kaiming He, Xiangyu Zhang, Shaoqing Ren, and Jian Sun. 2016.
\newblock Deep residual learning for image recognition.
\newblock In \emph{Proceedings of the IEEE conference on computer vision and
  pattern recognition}, pages 770--778.

\bibitem[{Howard and Ruder(2018)}]{Howard:2018acl}
Jeremy Howard and Sebastian Ruder. 2018.
\newblock Universal language model fine-tuning for text classification.
\newblock \emph{Proceedings of ACL}.

\bibitem[{Kiela(2017)}]{Kiela:17thesis}
Douwe Kiela. 2017.
\newblock \emph{{Deep Embodiment: Grounding Semantics in Perceptual
  Modalities}}.
\newblock Ph.D. thesis, University of Cambridge, Computer Laboratory.

\bibitem[{Kiela et~al.(2018)Kiela, Grave, Joulin, and Mikolov}]{Kiela:2018aaai}
Douwe Kiela, Edouard Grave, Armand Joulin, and Tomas Mikolov. 2018.
\newblock Efficient large-scale multi-modal classification.
\newblock In \emph{Thirty-Second AAAI Conference on Artificial Intelligence}.

\bibitem[{Kingma and Ba(2014)}]{Kingma:2014adam}
Diederik~P Kingma and Jimmy Ba. 2014.
\newblock Adam: A method for stochastic optimization.
\newblock \emph{arXiv preprint arXiv:1412.6980}.

\bibitem[{Kiros et~al.(2015)Kiros, Zhu, Salakhutdinov, Zemel, Urtasun,
  Torralba, and Fidler}]{Kiros:2015nips}
Ryan Kiros, Yukun Zhu, Ruslan~R Salakhutdinov, Richard Zemel, Raquel Urtasun,
  Antonio Torralba, and Sanja Fidler. 2015.
\newblock Skip-thought vectors.
\newblock In \emph{Advances in neural information processing systems}, pages
  3294--3302.

\bibitem[{Lazaridou et~al.(2015)Lazaridou, Pham, and
  Baroni}]{Lazaridou:2015skipgram}
Angeliki Lazaridou, Nghia~The Pham, and Marco Baroni. 2015.
\newblock Combining language and vision with a multimodal skip-gram model.
\newblock \emph{arXiv preprint arXiv:1501.02598}.

\bibitem[{Li et~al.(2019)Li, Yatskar, Yin, Hsieh, and
  Chang}]{Li:2019visualbert}
Liunian~Harold Li, Mark Yatskar, Da~Yin, Cho-Jui Hsieh, and Kai-Wei Chang.
  2019.
\newblock Visualbert: A simple and performant baseline for vision and language.
\newblock \emph{arXiv preprint arXiv:1908.03557}.

\bibitem[{Lu et~al.(2019)Lu, Batra, Parikh, and Lee}]{Lu:2019vilbert}
Jiasen Lu, Dhruv Batra, Devi Parikh, and Stefan Lee. 2019.
\newblock {ViLBERT: Pretraining Task-Agnostic Visiolinguistic Representations
  for Vision-and-Language Tasks}.
\newblock \emph{arXiv preprint arXiv:1908.02265}.

\bibitem[{Mikolov et~al.(2013)Mikolov, Sutskever, Chen, Corrado, and
  Dean}]{Mikolov:2013nips}
Tomas Mikolov, Ilya Sutskever, Kai Chen, Greg~S Corrado, and Jeff Dean. 2013.
\newblock Distributed representations of words and phrases and their
  compositionality.
\newblock In \emph{Advances in neural information processing systems}, pages
  3111--3119.

\bibitem[{Oquab et~al.(2014)Oquab, Bottou, Laptev, and Sivic}]{Oquab:2014cvpr}
Maxime Oquab, Leon Bottou, Ivan Laptev, and Josef Sivic. 2014.
\newblock Learning and transferring mid-level image representations using
  convolutional neural networks.
\newblock In \emph{Proceedings of the IEEE conference on computer vision and
  pattern recognition}, pages 1717--1724.

\bibitem[{Paszke et~al.(2017)Paszke, Gross, Chintala, and
  Chanan}]{Paszke:2017pytorch}
Adam Paszke, Sam Gross, Soumith Chintala, and Gregory Chanan. 2017.
\newblock {PyTorch: Tensors and dynamic neural networks in python with strong
  GPU acceleration}.
\newblock Technical report, PyTorch.

\bibitem[{Pennington et~al.(2014)Pennington, Socher, and
  Manning}]{Pennington:2014glove}
Jeffrey Pennington, Richard Socher, and Christopher Manning. 2014.
\newblock Glove: Global vectors for word representation.
\newblock In \emph{Proceedings of the 2014 conference on empirical methods in
  natural language processing (EMNLP)}, pages 1532--1543.

\bibitem[{Perez et~al.(2018)Perez, Strub, de~Vries, Dumoulin, and
  Courville}]{Perez:2018film}
Ethan Perez, Florian Strub, Harm de~Vries, Vincent Dumoulin, and Aaron
  Courville. 2018.
\newblock Film: Visual reasoning with a general conditioning layer.
\newblock In \emph{Proceedings of AAAI}.

\bibitem[{P{\'{e}}rez{-}R{\'{u}}a et~al.(2019)P{\'{e}}rez{-}R{\'{u}}a,
  Vielzeuf, Pateux, Baccouche, and Jurie}]{Perez-Rua:2019arxiv}
Juan{-}Manuel P{\'{e}}rez{-}R{\'{u}}a, Valentin Vielzeuf, St{\'{e}}phane
  Pateux, Moez Baccouche, and Fr{\'{e}}d{\'{e}}ric Jurie. 2019.
\newblock {MFAS:} multimodal fusion architecture search.
\newblock \emph{Proceedings of CVPR}, arxiv preprint 1903.06496.

\bibitem[{Peters et~al.(2018)Peters, Neumann, Iyyer, Gardner, Clark, Lee, and
  Zettlemoyer}]{Peters:2018naacl}
Matthew~E Peters, Mark Neumann, Mohit Iyyer, Matt Gardner, Christopher Clark,
  Kenton Lee, and Luke Zettlemoyer. 2018.
\newblock Deep contextualized word representations.
\newblock \emph{Proceedings of NAACL}.

\bibitem[{Phang et~al.(2018)Phang, F{\'{e}}vry, and Bowman}]{Phang2018sentence}
Jason Phang, Thibault F{\'{e}}vry, and Samuel~R. Bowman. 2018.
\newblock Sentence encoders on stilts: Supplementary training on intermediate
  labeled-data tasks.
\newblock \emph{CoRR}, abs/1811.01088.

\bibitem[{Radford et~al.(2018)Radford, Narasimhan, Salimans, and
  Sutskever}]{Radford:2018tr}
Alec Radford, Karthik Narasimhan, Tim Salimans, and Ilya Sutskever. 2018.
\newblock Improving language understanding by generative pre-training.
\newblock Technical report, OpenAI.

\bibitem[{Razavian et~al.(2014)Razavian, Azizpour, Sullivan, and
  Carlsson}]{Razavian:2014cvpr}
Ali~Sharif Razavian, Hossein Azizpour, Josephine Sullivan, and Stefan Carlsson.
  2014.
\newblock Cnn features off-the-shelf: an astounding baseline for recognition.
\newblock In \emph{Proceedings of the IEEE conference on computer vision and
  pattern recognition workshops}, pages 806--813.

\bibitem[{Socher et~al.(2013)Socher, Ganjoo, Manning, and Ng}]{Socher:2013nips}
Richard Socher, Milind Ganjoo, Christopher~D Manning, and Andrew Ng. 2013.
\newblock Zero-shot learning through cross-modal transfer.
\newblock In \emph{Advances in neural information processing systems}, pages
  935--943.

\bibitem[{Su et~al.(2019)Su, Zhu, Cao, Li, Lu, Wei, and Dai}]{Su:2019vlbert}
Weijie Su, Xizhou Zhu, Yue Cao, Bin Li, Lewei Lu, Furu Wei, and Jifeng Dai.
  2019.
\newblock Vl-bert: Pre-training of generic visual-linguistic representations.
\newblock \emph{arXiv preprint arXiv:1908.08530}.

\bibitem[{Sun et~al.(2019)Sun, Myers, Vondrick, Murphy, and
  Schmid}]{Sun:2019arxiv}
Chen Sun, Austin Myers, Carl Vondrick, Kevin Murphy, and Cordelia Schmid. 2019.
\newblock Videobert: {A} joint model for video and language representation
  learning.
\newblock \emph{arXiv preprint arXiv:1904.01766}.

\bibitem[{Tan and Bansal(2019)}]{Tan:2019lxmert}
Hao Tan and Mohit Bansal. 2019.
\newblock Lxmert: Learning cross-modality encoder representations from
  transformers.
\newblock \emph{arXiv preprint arXiv:1908.07490}.

\bibitem[{Tsai et~al.(2019)Tsai, Bai, Liang, Kolter, Morency, and
  Salakhutdinov}]{Tsai:2019acl}
Yao{-}Hung~Hubert Tsai, Shaojie Bai, Paul~Pu Liang, J.~Zico Kolter,
  Louis{-}Philippe Morency, and Ruslan Salakhutdinov. 2019.
\newblock Multimodal transformer for unaligned multimodal language sequences.
\newblock \emph{arxiv preprint 1906.00295}.

\bibitem[{Vaswani et~al.(2017)Vaswani, Shazeer, Parmar, Uszkoreit, Jones,
  Gomez, Kaiser, and Polosukhin}]{Vaswani:2017nips}
Ashish Vaswani, Noam Shazeer, Niki Parmar, Jakob Uszkoreit, Llion Jones,
  Aidan~N Gomez, {\L}ukasz Kaiser, and Illia Polosukhin. 2017.
\newblock Attention is all you need.
\newblock In \emph{Advances in neural information processing systems}, pages
  5998--6008.

\bibitem[{Vielzeuf et~al.(2018)Vielzeuf, Lechervy, Pateux, and
  Jurie}]{Vielzeuf:2018eccv}
Valentin Vielzeuf, Alexis Lechervy, Stephane Pateux, and Frederic Jurie. 2018.
\newblock Centralnet: a multilayer approach for multimodal fusion.
\newblock In \emph{The European Conference on Computer Vision (ECCV)
  Workshops}.

\bibitem[{Vu et~al.(2018)Vu, Greco, Erofeeva, Jafaritazehjan, Linders, Tanti,
  Testoni, Bernardi, and Gatt}]{Vu:2018vsnli}
Hoa~Trong Vu, Claudio Greco, Aliia Erofeeva, Somayeh Jafaritazehjan, Guido
  Linders, Marc Tanti, Alberto Testoni, Raffaella Bernardi, and Albert Gatt.
  2018.
\newblock Grounded textual entailment.
\newblock In \emph{Proceedings of COLING}, page 2354–2368.

\bibitem[{Wang et~al.(2019)Wang, Tran, and Feiszli}]{Wang:2019arxiv}
Weiyao Wang, Du~Tran, and Matt Feiszli. 2019.
\newblock What makes training multi-modal networks hard?
\newblock \emph{arXiv preprint arXiv:1905.12681}.

\bibitem[{Wang et~al.(2015)Wang, Kumar, Thome, Cord, and
  Precioso}]{Wang:2015food101}
Xin Wang, Devinder Kumar, Nicolas Thome, Matthieu Cord, and Frederic Precioso.
  2015.
\newblock Recipe recognition with large multimodal food dataset.
\newblock In \emph{2015 IEEE International Conference on Multimedia \& Expo
  Workshops (ICMEW)}, pages 1--6. IEEE.

\bibitem[{Weston et~al.(2011)Weston, Bengio, and Usunier}]{Weston:2011wsabie}
Jason Weston, Samy Bengio, and Nicolas Usunier. 2011.
\newblock Wsabie: Scaling up to large vocabulary image annotation.
\newblock In \emph{Twenty-Second International Joint Conference on Artificial
  Intelligence}.

\bibitem[{Young et~al.(2014)Young, Lai, Hodosh, and
  Hockenmaier}]{Young:2014flickr30k}
Peter Young, Alice Lai, Micah Hodosh, and Julia Hockenmaier. 2014.
\newblock From image descriptions to visual denotations: New similarity metrics
  for semantic inference over event descriptions.
\newblock \emph{Transactions of the Association for Computational Linguistics},
  2:67--78.

\bibitem[{Zadeh et~al.(2017)Zadeh, Chen, Poria, Cambria, and
  Morency}]{Zadeh2017tfn}
Amir Zadeh, Minghai Chen, Soujanya Poria, Erik Cambria, and Louis-Philippe
  Morency. 2017.
\newblock Tensor fusion network for multimodal sentiment analysis.
\newblock \emph{arXiv preprint arXiv:1707.07250}.

\end{thebibliography}
\bibliographystyle{acl_natbib}


\end{document}